# A Scoping Review of Large Language Model-Based Pedagogical Agents

Shan Li[1,2]
Email: shla22@lehigh.edu
ORCID: https://orcid/org/0000-0001-6001-1586

Juan Zheng[1]
Email: juz322@lehigh.edu
ORCID: https://orcid/org/0000-0002-4896-8198

[1]Department of Education and Human Services, College of Education, Lehigh University
[2]Department of Community and Global Health, College of Health, Lehigh University
Bethlehem, PA, USA, 18015

All correspondence should be addressed to Shan Li, HST Building 132, 124 E. Morton Street, Lehigh University, Bethlehem, PA 18015, USA, shla22@lehigh.edu

**Conflict of Interest**: The authors declare that they have no conflict of interests.

**Citation**:

Li, S., & Zheng, J. (2026). A scoping review of large language model-based pedagogical agents. *arXiv preprint.* https://arxiv.org/abs/2604.12253

# A Scoping Review of Large Language Model-Based Pedagogical Agents

## Abstract

This scoping review examines the emerging field of Large Language Model (LLM)-based pedagogical agents in educational settings. While traditional pedagogical agents have been extensively studied, the integration of LLMs represents a transformative advancement with unprecedented capabilities in natural language understanding, reasoning, and adaptation. Following PRISMA-ScR guidelines, we analyzed 52 studies across five major databases from November 2022 to January 2025. Our findings reveal diverse LLM-based agents spanning K-12, higher education, and informal learning contexts across multiple subject domains. We identified four key design dimensions characterizing these agents: interaction approach (reactive vs. proactive), domain scope (domain-specific vs. general-purpose), role complexity (single-role vs. multi-role), and system integration (standalone vs. integrated). Emerging trends include multi-agent systems that simulate naturalistic learning environments, virtual student simulation for agent evaluation, integration with immersive technologies, and combinations with learning analytics. We also discuss significant research gaps and ethical considerations regarding privacy, accuracy, and student autonomy. This review provides researchers and practitioners with a comprehensive understanding of LLM-based pedagogical agents while identifying crucial areas for future development in this rapidly evolving field.

**1. Introduction**

Pedagogical agents are virtual characters designed to facilitate learning through interaction with students (Dai et al., 2022; Schroeder et al., 2013; Zhang et al., 2024). Pedagogical agents have been extensively studied for decades. They evolved from simple text-based interfaces to animated and even embodied characters that support diverse learning activities through multimodal interactions (Casheekar et al., 2024; Dai et al., 2022; Yusuf et al., 2025). Traditional pedagogical agents often relied on rule-based systems and limited dialogue models, which restricted their ability for natural and adaptive interactions (Dai et al., 2022; Zhang et al., 2024). The recent emergence of Large Language Models (LLMs) represents a transformative advancement in artificial intelligence. LLMs demonstrate unprecedented capabilities in natural language understanding and generation, reasoning, and task completion across various domains (Wei et al., 2022).

Integrating LLMs into pedagogical agents could significantly enhance their capabilities and potential impact on education. LLM-based pedagogical agents can engage in more natural conversations, provide contextualized explanations, and adapt their teaching strategies based on student responses (Abdelghani et al., 2024; Hu et al., 2025; Liu & M'hiri, 2024; Shu et al., 2023; Wei et al., 2024). For example, Wei et al. (2024) demonstrated how LLM-powered pedagogical agents can seamlessly integrate with augmented reality for science education, while the agent *TeachTune* (Jin et al., 2024) showcased capabilities in simulating and evaluating pedagogical conversations with diverse student profiles. These advancements suggest that LLM-based pedagogical agents could revolutionize personalized learning and educational support.

While several review papers have examined related areas, there remains a critical gap in understanding specifically how LLMs are reshaping pedagogical agents. Zhang et al. (2024) provided a systematic review of pedagogical agent design for K-12 education, but their focus was primarily on traditional AI-based agents rather than those powered by LLMs. Wang et al. (2024) presented a comprehensive survey of LLM-based autonomous agents, but their scope encompassed general autonomous agents without specific attention to educational applications. Yusuf et al. (2025) examined pedagogical AI conversational agents in higher education, focusing on traditional AI approaches such as rule-based systems, machine learning algorithms, and natural language processing techniques, but did not address the emerging field of generative AI and LLM-based agents. These gaps in the literature underscore the need for a focused review of LLM-based pedagogical agents.

In this study, we conducted a scoping review to analyze the current landscape of LLM-based pedagogical agents. We chose a scoping review methodology due to the emerging nature of LLM-based pedagogical agents. Systematic reviews typically address specific questions about intervention effectiveness. In contrast, scoping reviews are ideal for examining the extent of research activity and identifying research gaps in developing fields (Munn et al., 2018; Peters et al., 2015). A scoping review allows us to comprehensively analyze the diverse implementations of pedagogical agents empowered by LLM. We aim to address the following research questions: (1) What is the current landscape of LLM-based pedagogical agents in terms of their applications in educational settings? (2) How are LLM-based pedagogical agents being designed and implemented to support teaching and learning? And (3) What emerging trends and future research directions can be identified in the field of LLM-based pedagogical agents? By addressing these three questions, this review would provide researchers and practitioners with a comprehensive understanding of the current state of LLM-based pedagogical agents while identifying crucial areas for future development in this rapidly evolving field.

## 2. Methods

This scoping review followed the PRISMA-ScR (Preferred Reporting Items for Systematic Reviews and Meta-Analyses Extension for Scoping Reviews) guidelines to ensure methodological rigor and transparent reporting (Tricco et al., 2018). The review specifically focused on LLM-based pedagogical agents in educational contexts.

*2.1 Search Strategy and Database Selection*

We conducted systematic searches across five major databases: ACM Digital Library, IEEE Xplore, ERIC (Education Resources Information Center), Web of Science, and arXiv. ACM Digital Library and IEEE Xplore were selected for their comprehensive coverage of computer science and technological implementations, while ERIC specializes in educational research. Web of Science was selected for its broad academic coverage at the intersection of education and AI across multiple disciplines. Given the rapid evolution of LLM technology, we also included arXiv to capture recent developments that may not yet have appeared in traditional peer-reviewed venues.

The following search syntax was used as a base template and adapted according to each database's specific syntax requirements: (“large language model*” OR “LLM*” OR “ChatGPT” OR “GPT*” OR “PaLM” OR “Claude” OR “Gemini” OR “LLaMA” OR “Mistral*” OR “generative AI” OR “generative artificial intelligence”) AND (“agent*” OR “tutor*” OR “mentor*” OR “companion*” OR “assistant*”) AND (“education*” OR “learning” OR “teaching” OR “instruction*” OR “classroom*” OR “pedagog*” OR “academic*”). For LLM technology terms, the search string combines general terminology (“large language model*”, “LLM*”, “generative AI”, “generative artificial intelligence”) with specific model names (“ChatGPT”, “GPT*”, “PaLM”, “Claude”, “Gemini”, “LLaMA”, “Mistral*”) to ensure comprehensive coverage, as authors often use either broad terms or reference specific LLM implementations in their work. Regarding pedagogical agent terms, “agent*” was chosen as the primary search term as it encompasses various agent-specific phrases (e.g., pedagogical agents, teaching agents, tutoring agents, educational agents, conversational agents), while additional terms “tutor*”, “mentor*”, “companion*” and “assistant*” were included separately because they represent distinct educational roles that might not always be described using agent terminology in the literature. The combination of those terms captures the intersection of LLM, agents, and teaching/learning contexts, with asterisks (*) enabling truncated searching to include plural forms and variations.

For each database, we conducted searches within the abstracts rather than full text or all metadata fields, as abstract-level searching typically yields more precise and relevant results while reducing false positives that might occur when search terms appear only peripherally in the full text (Gusenbauer & Haddaway, 2020). The search covered the period from November 1, 2022 (coinciding with ChatGPT’s release) through January 30, 2025. This timeframe was chosen to capture the emergence and evolution of LLM-based pedagogical agents following the widespread availability of powerful language models. Search results were exported and managed using Mendeley reference management software.

*2.2 Inclusion and Exclusion Criteria*

Studies were included if they met the following criteria: (1) focused on pedagogical agents specifically powered by large language models, (2) demonstrated clear educational applications or learning support functions, and (3) appeared in English-language publications. The review encompassed conference papers, journals, review papers, and early access articles. For review papers, we focused on extracting key insights that could inform our analysis of LLM-based pedagogical agents, rather than conducting a critical review of reviews, as this would have shifted focus away from our primary goal of understanding the design and implementation of LLM-based agents in education.

We excluded studies if they (1) used traditional rule-based or non-LLM AI approaches for pedagogical agents, or (2) lacked substantial integration of LLMs into pedagogical agent design and implementation. Regarding the second criterion, this included articles that merely mentioned LLMs without educational context (e.g., papers focused on technical aspects of language models), papers examining LLM-based agents in non-educational domains (e.g., customer service, healthcare), theoretical or position papers without actual implementation of LLM-based agents, studies focused solely on perceptions of generative AI in education without specific examination of LLM-based pedagogical agents, and articles discussing only educational policy, ethics, or governance of pedagogical AI agents. We also excluded commentaries, opinion pieces, and letters to editors, as these typically do not provide detailed technical implementations or empirical evidence regarding how LLMs are being integrated into pedagogical agents. Additionally, we excluded books and book chapters.

*2.3 Screening Process*

The screening process was conducted in multiple phases (see Figure 1). In the identification phase, our database searches yielded a total of 1,254 records across the five selected databases (ACM Digital Library, IEEE Xplore, ERIC, Web of Science, and arXiv). After removing 203 duplicate records and 6 book chapters, 1045 records remained for screening.

In the screening phase, the first author, who has expertise in educational technology and AI in education, independently reviewed all abstracts based on the predefined exclusion criteria. This initial screening focused on identifying studies that explicitly discussed LLM-based pedagogical agents in educational contexts. Records were excluded if they focused solely on technical aspects of language models without educational applications, dealt with non-LLM approaches, or were purely theoretical without implementation details. This process resulted in 160 potentially relevant studies for full-text review.

During the eligibility phase, two researchers collaboratively assessed the full texts of these 160 studies to determine if they met the inclusion criteria. The researchers discussed the inclusion criteria and resolved any disagreements through consensus. Studies were included if they focused on pedagogical agents specifically powered by LLMs and demonstrated clear educational applications or learning support functions. Through this collaborative review process, 108 studies were excluded for various reasons, including insufficient focus on LLM integration, lack of pedagogical agent implementation, or inadequate context.

The final selection resulted in 52 studies that met the inclusion criteria and were included in the review. These studies represented a diverse range of LLM-based pedagogical agent implementations across various educational contexts and subject domains. The relatively small number of included studies reflects both the emerging nature of this field and our stringent inclusion criteria focusing specifically on LLM-based pedagogical agents rather than general AI applications in education.

Figure 1 *Literature Search and Selection Process for LLM-based Pedagogical Agents*

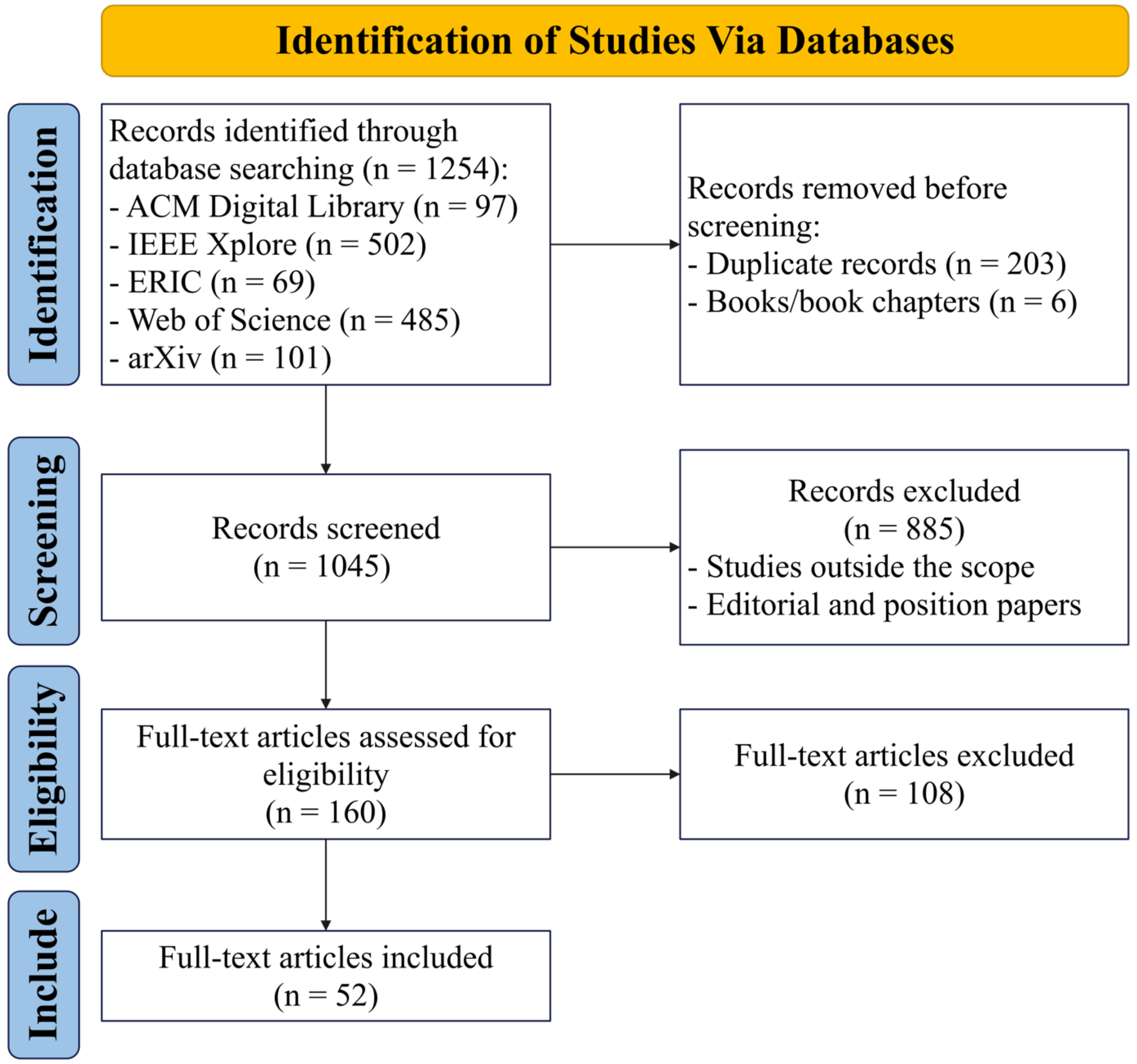

*2.4 Data Extraction and Synthesis*

Data extraction was guided by our three research questions examining the current landscape, design/implementation approaches, and emerging trends of LLM-based pedagogical agents in educational settings. For the first research question on the current landscape, we extracted information about educational contexts (e.g., K-12, higher education, informal learning), subject domains (e.g., science, language learning, programming), and intended learning outcomes of LLM-based pedagogical agents. We also documented the specific roles and functions of these agents. For the second research question on design and implementation approaches, we focused on how these agents were being conceptualized and deployed to support teaching and learning. Through detailed review of representative papers such as Casheekar et al. (2024), Wei et al. (2024), Wu et al. (2023), and Yu et al. (2024), we identified four key dimensions that characterize LLM-based pedagogical agents: interaction approaches (reactive vs. proactive), domain scope (domain-specific vs. general-purpose), role complexity (single-role vs. multi-role), and system integration (standalone vs. integrated). These dimensions formed an analytical framework for extracting and categorizing data on how agents were designed and implemented across the literature. For the third question on emerging trends, we analyzed patterns across implementations from multiple studies to identify convergent developments and organized them into thematic categories. Given the nature of a scoping review, our goal was not to exhaustively discuss every article for each research question, as would be expected in a systematic review (Munn et al., 2018). Instead, we focused on mapping the current landscape of LLM-based pedagogical agents and identifying key patterns and trajectories in how this field is evolving, using representative examples to illustrate significant developments and trends.

## 3. Results

### *3.1 What is the current landscape of LLM-based pedagogical agents in terms of their applications in educational settings?*

The current landscape of LLM-based pedagogical agents is characterized by a wide range of applications across various educational settings, as evidenced by the diverse systems described in the Appendix. These applications span different educational levels, subject domains, and learning contexts. Table 1 presents representative examples of these applications, which illustrates the broad spectrum of ways LLM-based pedagogical agents are being utilized in educational environments.

**Table 1** *LLM-Based Pedagogical Agents by Application Categories*

| **Primary User** | **Application Category** | **Contexts and Example Agents** |
|---|---|---|
| **Student-Facing** | Subject-Specific Education | • Elementary science/astronomy education:<br>*GPA (Wei et al., 2024)*<br>• High school science education:<br>*ScienceChat (Liu, Lu et al., 2024)*<br>• Programming education:<br>*Python Pedagogical Agent (Chiang et al., 2024)*<br>• English language learning:<br>*ELLMA-T (Pan et al., 2024)*<br>• Anatomical sciences education:<br>*AnatomyGPT (Collins et al., 2024)*<br>• Business education:<br>*ChEdBot (Martin et al., 2024)* |
| | General Educational Support | • Divergent question-asking skills:<br>*Kids Ask (Abdelghani et al., 2024)*<br>• Collaborative learning facilitation:<br>*PeerGPT (Liu, Yao et al., 2024)* |
| | Multiagent Learning Environments | • Interactive virtual classmate:<br>*ClassMeta (Liu, Zhu et al., 2024)*<br>*EduAgent (Xu et al., 2024)*<br>• Multi-agent classroom environment:<br>*MAIC (Yu et al., 2024)* |
| **Teacher-Facing** | Instructional Design Support | • Pedagogical agent design and evaluation:<br>*TeachTune (Jin et al., 2024)*<br>• Physics simulation customization:<br>*SimPal (Farhana et al., 2024)* |
| | Classroom Management and Facilitation | • Specialized sub-agents for flipped classrooms:<br>*CVTutor (Teng et al., 2024)*<br>• Naturally interactable teaching assistant:<br>*NivTA (Jia et al., 2024)* |
| | Assessment and Feedback | • Multiple-choice question generation:<br>*MCQ Generation System (Wang et al., 2024)*<br>• Open-ended response evaluation:<br>*FreeText (Matelsky et al., 2023)* |

*3.1.1 Student-Facing Agents*

LLM-based pedagogical agents have been applied to diverse subject domains across educational levels, such as elementary and secondary science (Liu, Lu et al., 2024; Wei et al., 2024), programming (Bassner et al., 2024; Chiang et al., 2024; Glynn et al., 2024; Martínez-Araneda et al., 2023), literacy (Pan et al., 2024), anatomical sciences education (Collins et al., 2024), and business education (Martin et al., 2024). For instance, Wei et al. (2024) implemented a Generative AI-empowered Pedagogical Agent (GPA) system using the 5E learning model to teach astronomy. The system includes three different roles, i.e., Instructor, Guide, and Conversation GPA, that deliver content, facilitate learning activities, and provide contextual explanations. Similarly, Liu, Lu et al. (2024) developed *ScienceChat*, an AI-based tutoring platform for high school science education, which supports both chat-based and quiz-based learning. *ScienceChat* allows students to engage in open-ended conversations with an AI agent, providing personalized feedback and guiding them through complex scientific concepts. The platform also generates multiple-choice questions based on the video content to reinforce student learning. In programming education, Chiang et al. (2024) introduced a Python Pedagogical Agent powered by ChatGPT to transition from outcome-oriented to process-outcome-balanced programming education practices. Specifically, they proposed a ChatGPT-supported experiential learning cycle combined with multimodal learning portfolios, where students engage in hands-on programming tasks, reflect on their learning experiences, and receive personalized feedback from the AI agent. Language learning has also benefited from LLM-based agents. Pan et al. (2024) developed *ELLMA-T*, an embodied LLM agent designed to support English language learning in social Virtual Reality (VR) environments. *ELLMA-T* leverages GPT-4 to simulate realistic, context-specific role-play scenarios within the VR environment, which provides learners with immersive, real-life conversation experiences. In health science education, Collins et al. (2024) developed *AnatomyGPT* to provide medical students with detailed, domain-specific guidance. Furthermore, the application of pedagogical AI agents extends to business education. For instance, Martin et al. (2024) developed *ChEdBot* for business education, which allows students to interview AI-powered domain experts and develop requirements-gathering skills.

Beyond specific subject areas, LLM-based pedagogical agents address general educational needs, such as fostering critical thinking, curiosity, inquiry, and collaborative skills across disciplines. For instance, *Kids Ask* is a conversational agent designed to train students in divergent question-asking (Abdelghani et al., 2024), a skill that transcends subject boundaries and promotes curiosity-driven learning. By leveraging GPT-3, *Kids Ask* provides students with linguistic and semantic cues to help them formulate open-ended, exploratory questions, encouraging them to think beyond rote memorization and engage deeply with the material. Similarly, *PeerGPT*, as explored by Liu, Yao et al. (2024), demonstrates the potential of LLM-based peer agents in children's collaborative learning. *PeerGPT* functions as both a team moderator and a participant. As a team moderator, it manages discussions by structuring tasks, guiding conversations, and summarizing ideas. As a participant, it contributes knowledge, stimulates creative thinking, and responds to children's inquiries. The agent adapts its role based on the needs of the group, which enhances the overall quality of collaborative learning.

Lastly, a significant advancement is the development of multiagent learning environments that create more naturalistic and engaging learning spaces. For instance, *ClassMeta* (Liu, Zhu et al., 2024) promotes participation in virtual classrooms by simulating active student behaviors through GPT-powered agents. The agents display realistic behaviors such as note-taking, question-asking, and contributing to discussions to establish positive behavioral norms in the virtual classroom environment. Similarly, *EduAgent* (Xu et al., 2024) creates a robust simulation framework that replicates diverse student learning behaviors in online education contexts. Through the integration of cognitive theories and large language models, *EduAgent* produces virtual students who exhibit naturalistic gaze patterns, motor behaviors, emotional states and learning outcomes that mirror real student interactions. The *MAIC* (Massive AI-empowered Course) system (Yu et al., 2024) transforms traditional MOOCs into adaptive learning environments by employing multiple specialized agents including a teacher agent, teaching assistants, and customizable peer agents who collaboratively create an interactive classroom environment. Each agent serves distinct pedagogical functions: the teacher agent delivers content and guides discussions, teaching assistants provide support and maintain classroom flow, while peer agents engage in discussions and demonstrate model learning behaviors. The three systems collectively demonstrate how LLM-powered agents can create authentic educational simulations that enhance students' learning experience.

#### *3.1.2 Teacher-Facing Agents*

In addition to direct student interaction, LLM-based pedagogical agents are increasingly supporting teacher instructional design processes (Farhana et al., 2024; Jin et al., 2024; Makharia et al., 2024). For instance, *TeachTune* is a tool that allows teachers to design and review pedagogical conversational agents (PCAs) by simulating conversations with diverse student profiles (Jin et al., 2024). *TeachTune* leverages LLMs to create simulated students with varying knowledge levels and traits, enabling teachers to test the adaptability of their PCAs across different learning scenarios. This approach reduces the manual effort required for multi-turn conversations and helps teachers identify gaps in their instructional designs. As another example, Farhana et al. (2024) introduced *SimPal*, a meta-conversational framework designed to assist K-12 physics teachers in adopting and customizing existing physics simulations. Through natural language conversations, *SimPal* helps teachers define their desired learning outcomes and generates symbolic representations of these goals, which can then be used to customize the AI agent's prompts.

Moreover, specialized agents have been developed to assist with classroom management and facilitation. For instance, Teng et al. (2024) developed *CVTutor*, a system with specialized sub-agents for flipped classrooms. These agents handle different aspects of course management, including answering student questions, grading assignments, and ensuring ethical compliance. *NivTA* (Jia et al., 2024) serves as a naturally interactable teaching assistant that uses multiple modalities including speech, gesture, gaze, and spatial awareness to provide more human-like educational interactions. This system demonstrates how LLM-based agents can be embodied in ways that enhance their presence and effectiveness in educational settings.

Assessment and feedback represent another crucial area where LLM-based pedagogical agents are making significant contributions. The *FreeText* system, developed by Matelsky et al. (2023), offers an automated solution for evaluating open-ended student responses. This system addresses the challenge of providing timely, consistent feedback in large educational settings where personalized instructor attention is time-intensive. *FreeText* employs instructor-defined grading criteria and feedback templates, allowing for customized assessment parameters while maintaining pedagogical standards. Wang et al.'s (2024) Multi-Agent MCQ (Multiple-Choice Question) Generation System takes a different approach, focusing on creating high-quality MCQs for AI literacy assessment. The system employs a Generator Agent to produce initial questions, which are subsequently evaluated by two independent critique agents: a Language Critique Agent and an Item-Writing Flaw Critique Agent. These agents assess the questions for readability, grade-level appropriateness, and potential compositional issues. The system iteratively refines the questions based on these evaluations, with a Supervisor Agent ultimately rendering final approval. This collaborative approach ensures that generated questions align with learning objectives and meet pedagogical standards. Together, the two examples showcase the potential for LLM-based pedagogical agents to address different aspects of educational assessment, from individual feedback to test item generation.

*3.2 How are LLM-based pedagogical agents being designed and implemented to support teaching and learning?*

Through our review, we found that LLM-based pedagogical agents can be characterized along four key dimensions: interaction approach, domain scope, role complexity, and system integration (see Table 2). These dimensions reflect fundamental design decisions that shape how agents engage with learners, what knowledge they possess, what pedagogical functions they serve, and how they connect with other educational tools and environments. Rather than being mutually exclusive categories, these dimensions often intersect and combine in various ways to create effective educational agents.

**Table 2** *Types of LLM-Based Pedagogical Agents*

| Dimension | Type | Characteristics |
|---|---|---|
| Interaction Approach | Reactive | • Responds to student-initiated queries and actions<br>• Provides support when explicitly requested |
| | Proactive | • Actively monitors student behavior<br>• Initiates interventions based on learning progress |
| Domain Scope | Domain-specific | • Deep knowledge in particular subject areas<br>• Integration with subject-specific tools |
| | General-purpose | • Flexible application across contexts<br>• Generic learning support capabilities |
| Role Complexity | Single-role | • Focus on one primary role and function<br>• Streamlined interaction model |
| | Multi-role | • Multiple pedagogical roles and functions<br>• Distributed support capabilities |
| System Integration | Standalone | • Self-contained functionality<br>• Direct user interaction |
| | Integrated | • Works with other educational tools or environments<br>• Ecosystem connectivity |

*3.2.1 Interaction Approach: Reactive vs. Proactive Agents*

The interaction approach of pedagogical agents determines how they engage with users and initiate educational interactions. This dimension encompasses two primary approaches: reactive and proactive, with many systems incorporating elements of both in hybrid implementations (Jin et al., 2024; Neyem et al., 2024; Xu et al., 2024).

Reactive agents respond primarily to student-initiated queries and actions, providing support when explicitly requested. For instance, Wei et al. (2024) adopted a reactive approach in their implementation of a Generative AI-empowered Pedagogical Agent (GPA) within the 5E learning model framework. Their GPA waits for student initiation before providing support across all phases of the learning process. During the Engagement phase, the GPA introduces topics and responds to initial student curiosity about science concepts. In the Exploration phase, it answers queries as students investigate phenomena. Throughout the Explanation phase, it provides clarification and structured responses to student questions. During Elaboration, students can prompt the GPA for guided responses that help extend their understanding. In the Evaluation phase, students communicate with the GPA to assess their knowledge mastery and receive instant feedback. The *CodeAid* system exemplifies another primarily reactive approach, as it responds to student queries about programming concepts or code issues (Kazemitabaar et al., 2024). Students select a feature (such as General Question, Help Fix Code, or Explain Code), and *CodeAid* generates responses based on the student's input. As another example, the AI agent *TutorBot+* provides feedback when students request assistance during programming tasks (Martínez-Araneda et al., 2023). Similarly, Lee's (2024) statistics tutor operates reactively, allowing students to pose questions in natural language about statistical concepts and methods.

In contrast, proactive agents actively monitor student behavior and learning patterns to initiate timely interventions. These agents employ monitoring mechanisms to identify potential learning difficulties, engagement drops, or opportunities for deeper learning (Jia et al., 2024; Meng et al., 2024; Zhang et al., 2024). In Zhang et al.'s (2024) research, teacher agents proactively guide classroom discussions, pose questions to stimulate engagement, and dynamically adjust teaching strategies based on real-time assessment of student comprehension. The mathematical tutoring agent platform developed by Meng et al. (2024) incorporates proactivity by breaking down complex mathematical word problems into manageable sub-problems and gradually guiding students through solution steps. Moreover, *NivTA* (Naturally Interactable Virtual Teaching Assistant) exemplifies spatial awareness in its interaction model (Jia et al., 2024). The agent can greet users as they approach the learning environment and initiate conversations based on proximity (Jia et al., 2024).

Several systems effectively integrate both reactive and proactive elements to create comprehensive educational experiences. For instance, Wang et al. (2024) developed a pedagogical framework that not only responds to student knowledge queries but also proactively generates AI literacy assessment questions tailored to learning objectives. Xu et al. (2024) presented *EduAgent*, which adapts its teaching approach by monitoring cognitive states, motivation levels, and academic stress while maintaining responsive dialogue capabilities. *TeachTune* enables teachers to test specific conversational scenarios by interacting with the agent and receiving targeted responses (Jin et al., 2024). Additionally, *TeachTune* takes proactive steps to generate classroom interactions that challenge and support learners. Lastly, Neyem et al. (2024) developed an AI Knowledge Assistant that responds to student queries and proactively suggests relevant “lessons learned” from previous projects based on the context of students’ work in a Kanban project management system. As Lee, Shin et al. (2024) noted in their study of LLM-based task-oriented dialogue chatbots, these systems create a more engaging learning experience than previous chatbot iterations.

*3.2.2 Domain Scope: Domain-specific vs. General-purpose Agents*

The domain scope of pedagogical agents reflects their subject matter expertise and application range, with significant implications for how they support different learning contexts and educational objectives. This dimension encompasses two main approaches: domain-specific and general-purpose.

Domain-specific agents are designed with deep knowledge of a particular subject area or educational domain. These specialized agents offer targeted support for learning within their field of expertise, providing highly relevant content, examples, and assistance (Bassner et al., 2024; Collins et al., 2024). For instance, Wei et al. (2024) developed a domain-specific

GPA for elementary science education, focusing particularly on astronomy concepts like the solar system, eclipses, and celestial phenomena. Similarly, Liu, Lu et al. (2024) created *ScienceChat*, a platform specifically designed for high school science education that uses retrieval-augmented generation to ground its responses in validated scientific content from educational videos. The *Iris* agent developed by Bassner et al. (2024) focused exclusively on computer science education. For medical education, Collins et al. (2024) developed *AnatomyGPT* specifically for anatomical sciences education. The *AnatomyGPT* was configured with open-source textbooks on anatomy, physiology, and histology as knowledge sources, allowing it to provide detailed, domain-specific guidance to medical students. Moreover, as Hu et al. (2025) demonstrated with *CollaBot*, domain-specific knowledge enhances the ability to provide targeted cognitive scaffolding in collaborative writing.

General-purpose agents, conversely, are designed to support learning across multiple subjects and contexts. These agents emphasize flexibility and adaptability over depth in any particular domain, making them suitable for interdisciplinary learning and for addressing diverse student needs within a single platform. For instance, the *Kids Ask* agent was designed to train students in divergent question-asking skills applicable across different subjects (Abdelghani et al., 2024). As another example, *VidBot* focuses on video content analysis for educational purposes, generating mind maps, knowledge graphs, and facilitating intelligent tutoring to video-based learning (Xie et al., 2024). The Educational Copilot discussed by Bekeš and Galzina (2023) represents another example, enabling teachers to create unit plans, presentations, quizzes, and other materials across various subjects.

Researchers have also incorporated elements of both domain-specific and general-purpose approaches to design LLM-based pedagogical agents. Yu et al.'s (2024) MAIC (Massive AI-empowered Course) represents a hybrid approach, where the primary teaching agent possesses deep domain knowledge of the course content, while teaching assistants and peer agents provide more general support for learning strategies and engagement. Similarly, *NivTA* provides specialized knowledge about educational concepts visualized in its knowledge graph while offering general knowledge to engage in diverse conversations with students (Jia et al., 2024). The choice between domain-specific and general-purpose agents often depends on educational goals and implementation contexts. The decision ultimately reflects a balance between depth of expertise and breadth of application, with each approach offering distinct advantages for different educational scenarios.

*3.2.3 Role Complexity: Single-role vs. Multi-role Agents*

Recent advances in LLMs have significantly expanded the potential for role complexity in pedagogical agents, allowing them to effectively simulate diverse pedagogical approaches and adapt their communication styles to different educational roles (Casheekar et al., 2024; Dai et al., 2022; García-Méndez et al., 2024; Liu & M'hiri, 2024; Zhang et al., 2024). In general, the role complexity of LLM-based agents ranges from single-role specialists focusing on one primary pedagogical function to multi-role systems that embody or coordinate multiple educational roles.

Single-role agents focus on executing one primary pedagogical function effectively, whether as tutors, coaches, peers, evaluators, or information providers. For instance, Matelsky et al.'s (2023) *FreeText* system maintains a singular focus on assessment and feedback for open-ended student responses. Some agents like *Iris* specialize exclusively in the tutor role, providing calibrated assistance for programming exercises through subtle hints and counter-questions rather than complete solutions (Bassner et al., 2024). Single-role agents offer certain benefits, including simplicity and clarity of purpose; as Martin et al. (2024) noted with *ChEdBot*, a well-defined single role with clear responsibilities can lead to more focused and effective interactions within a specific domain.

Multi-role agents embody multiple pedagogical functions within a single system, transitioning between different roles based on learning contexts, student needs, or explicit directives. A typical example is Wei et al.'s (2024) three-role GPA system, which distributes different pedagogical functions across specialized agent personas: the Instructor GPA functions as a knowledge presenter, the Guide GPA serves as a task facilitator, and the Conversation GPA responds to student queries with contextual explanations. Similarly, Yu et al. (2024) developed MAIC, employing multiple agents with distinct roles: a teacher agent for delivering content and guiding classroom activities, teaching assistants for supplementary explanations, and peer agents for modeling ideal student behaviors. *NivTA* demonstrates this versatility by functioning as both a tutor for educational concepts and a lecture reenactment system (Jia et al., 2024). As noted by Zhang et al. (2024), distributing pedagogical functions across multiple agents allows each to specialize in particular aspects of the educational process, potentially creating more engaging and dynamic learning environments.

*3.2.4 System Integration: Standalone vs. Integrated Agents*

The system integration dimension examines how LLM-based pedagogical agents exist within broader educational technology ecosystems. This dimension ranges from standalone applications that function independently to integrated components that work in conjunction with other educational tools and platforms (Casheekar et al., 2024; Dai et al., 2022; García-Méndez et al., 2024; Kim & Baylor, 2016; Zhang et al., 2024).

Standalone agents operate independently, providing self-contained educational experiences without requiring connection to other learning systems. For instance, the Python Pedagogical Agent functions entirely within ChatGPT's interface, where students engage in natural dialogue, submit code, and receive feedback within a single chat window (Chiang et al., 2024). Similarly, the task-oriented dialogue chatbots examined by Lee, Shin et al. (2024) serve as independent conversational partners for language learning without external platform integration. Standalone agents offer several advantages, including reduced technical complexity, clear interaction channels, and simplified deployment. However, as pointed out by Teng et al. (2024) and Wang et al. (2024), standalone agents may lack access to specialized domain tools that could enhance learning experiences and miss out on the rich contextual data available in existing learning systems.

In contrast, integrated agents work in conjunction with other educational technologies, forming part of a broader learning ecosystem. Wei et al.'s (2024) GPA system is a good example that works seamlessly with Unity's AR platform, the Vuforia plugin, and ChatGPT's API to facilitate AR-based science lessons. As another example, the *Iris* agent is based on the Artemis learning platform, allowing it to access problem statements, student code, and automated feedback for context-aware assistance (Bassner et al., 2024). Integrated agents offer significant benefits, including enhanced personalization through multi-source data, augmentation of existing educational tools, and rich contextual awareness (Bran et al., 2023; Yusuf et al., 2025). For instance, *VidBot* integrates with video content systems, enabling analysis of video transcripts and personalized assistance based on playback patterns (Xie et al., 2024). However, integrated approaches present challenges, including complex architectural requirements, greater technical expertise for implementation, and potential data privacy concerns due to broader system access.

*3.3 What emerging trends and future research directions can be identified in the field of LLM-based pedagogical agents?*

Our review reveals several emerging trends and future research directions in the field of LLM-based pedagogical agents. Given the nascent and rapidly evolving nature of this field (Wang et al., 2024), a scoping review provides a snapshot of current developments. While we cannot claim to capture all emerging trends with the same rigor as a systematic review might offer for a more established field, this approach allows us to identify important patterns across diverse implementations (Munn et al., 2018). The themes presented here emerged from our analysis of the literature and represent recurring directions that multiple research teams are pursuing independently.

A prominent trend is the development of multi-agent systems to create more immersive and realistic educational environments. Yu et al. (2024) designed the MAIC (Massive AI-empowered Course) system, where multiple agents with distinct roles collaborate to create a dynamic classroom experience. Similarly, Zhang et al. (2024) developed *SimClass*, a multi-agent classroom simulation framework where teacher agents, teaching assistants, and student agents interact to facilitate learning. This shift toward complex multi-agent systems reflects a recognition that effective educational environments often involve multiple pedagogical roles and social dynamics that a single agent cannot replicate. Future research could explore how to optimize coordination between multiple agents and ensure coherent educational narratives across agent interactions.

Additionally, student simulation has emerged as a significant innovation for evaluating and improving pedagogical agents. For instance, Xu et al. (2024) introduced *EduAgent*, a framework for generating virtual students with realistic learning behaviors to test educational systems before deployment with real students. Similarly, Jin et al. (2024) developed *TeachTune*, which helps teachers review and improve their pedagogical conversational agents by simulating interactions with diverse student profiles. The ability to simulate diverse student profiles with varying knowledge levels, learning styles, and behavioral patterns provides a powerful tool for optimizing pedagogical agent design.

Another significant trend is the integration of LLM-based agents with immersive technologies like augmented and virtual reality. Wei et al. (2024) pioneered this approach by embedding their GPA system within an AR environment for science education, allowing students to interact with 3D models while receiving guidance from virtual tutors. Likewise,

Pan et al. (2024) developed *ELLMA-T*, an embodied LLM agent for language learning in social VR environments. These implementations suggest that LLMs can significantly enhance immersive learning experiences by providing contextually relevant guidance and personalized support. Future research could investigate how to optimize LLM-agent embodiment in virtual spaces, develop multimodal interaction capabilities that combine verbal and non-verbal communication, and design pedagogical approaches specifically suited to immersive learning contexts (Casheekar et al., 2024; Gao et al., 2023).

Moreover, researchers are developing pedagogical agents that are more naturally interactable and adaptive to student needs. For instance, *NivTA* incorporates interaction modalities beyond text, including speech, gesture, gaze, and spatial awareness (Jia et al., 2024). As these types of agents continue to evolve, they may increasingly approximate human-like educational interactions, potentially enhancing student engagement and learning outcomes. The AI tutor, developed by Makharia et al. (2024), combines prompt engineering with Deep Knowledge Tracing to provide assistance calibrated to a student's proficiency level. Future research will likely expand on these adaptive capabilities, developing more nuanced models of student knowledge and response strategies.

The combination of pedagogical agents with learning analytics represents another promising future direction (Lee et al., 2020; Mangaroska & Giannakos, 2018; Xie et al., 2024). For instance, the *VidBot* agent combines playback traffic analysis with intelligent tutoring capabilities (Xie et al., 2024). This integration demonstrates the significant potential of merging learning analytics with AI-driven pedagogical support, allowing for more personalized and contextually relevant guidance based on actual student viewing patterns (Xie et al., 2024). Future research might explore more analytics integration, using behavioral data to inform agent responses and adaptively optimize learning experiences. By leveraging data from multiple sources, including student interactions, content engagement metrics, and performance assessments, LLM-based agents could provide increasingly personalized and effective educational support.

Furthermore, several researchers have identified the need for better evaluation frameworks specifically designed for LLM-based pedagogical agents. Liu and M'hiri (2024) highlighted that traditional evaluation methods may not adequately capture the unique capabilities and limitations of these systems. Teng et al. (2024) employed both technical and human evaluations in their assessment of *CVTutor*, but noted the challenges in standardizing evaluation approaches across different educational contexts. Future research should focus on developing robust frameworks for assessing both the technical performance and pedagogical effectiveness of LLM-based agents. Future studies might employ experimental designs to compare learning outcomes between traditional instruction and agent-mediated approaches, providing stronger evidence for the educational value of these systems. Longitudinal research examining the sustained impact of these agents on learning trajectories would be particularly valuable. Equally important would be studies exploring how these technologies might affect different student populations and learning contexts.

The field of LLM-based pedagogical agents is rapidly evolving. Future research that addresses these directions has the potential to significantly advance the effectiveness and impact of AI-powered educational support systems by creating more natural, adaptive, and pedagogically sound learning experiences across diverse educational contexts.

## 4. Discussion

### *4.1 Synthesis of Key Findings*

Regarding our first research question on the current landscape of LLM-based pedagogical agents, our review reveals a diverse ecosystem spanning multiple educational contexts and subject domains. These agents are being deployed across various subject-specific applications including elementary science education (Wei et al., 2024), programming education (Bassner et al., 2024; Glynn et al., 2024; Kazemitabaar et al., 2024; Martínez-Araneda et al., 2023), literacy (Pan et al., 2024), health science education (Collins et al., 2024), and business education (Martin et al., 2024). Beyond subject-specific implementations, LLM-based agents also address cross-disciplinary educational needs such as fostering critical thinking, inquiry skills, metacognition, and collaborative learning (Abdelghani et al., 2024; Li et al., 2025; Liu, Yao et al., 2024). The landscape is further characterized by several distinctive capabilities that differentiate these systems from previous educational technologies: adaptive personalization of learning experiences (as demonstrated by the *MAIC* system) (Yu et al., 2024), interactive agent-based simulations that model realistic educational interactions (exemplified by *ClassMeta* and *EduAgent*) (Liu, Zhu et al., 2024; Xu et al., 2024), and nuanced assessment approaches that provide tailored feedback (illustrated by *FreeText* and the MCQ Generation System) (Matelsky et al., 2023; Wang et al., 2024). These capabilities represent a significant advancement over previous generations of pedagogical agents, as LLM-based agents demonstrate unprecedented natural language understanding and generation (Dai et al., 2022; Lee, 2024; Wang et al., 2024).

For our second research question on design and implementation approaches, we proposed an analytical framework, which includes dimensions of interaction approach, domain scope, role complexity, and system integration. These dimensions reveal important patterns in how researchers are conceptualizing and implementing these agents. The four dimensions are not isolated but rather intersect in complex ways that shape agent functionality. For instance, domain-specific agents like *Iris* (Bassner et al., 2024) often achieve greater depth in their educational support but at the cost of flexibility across contexts, while general-purpose agents like *Kids Ask* (Abdelghani et al., 2024) offer broader applicability but may lack specialized knowledge. Similarly, we find important relationships between role complexity and interaction approach, with multi-role systems like Wei et al.'s 92024) *GPA* or Yu et al.'s (2024) *MAIC* typically incorporating both reactive and proactive elements distributed across specialized agent personas. The integration dimension reveals a significant shift in how LLM-based pedagogical agents connect with existing educational tools as they offer more flexible and seamless incorporation through their enhanced natural language understanding capabilities. Beyond these dimensional patterns, our review highlights the transformative impact of prompt engineering as a design approach. This represents a fundamentally different paradigm for pedagogical agent design that prioritizes natural language instruction over conventional programming (Dai et al., 2022).

While section 3.3 identified emerging trends in the field of LLM-based pedagogical agents, our analysis reveals several gaps that warrant more research. First, there is insufficient attention to teacher perspectives and integration practices in actual classroom settings. While some studies like *TeachTune* (Jin et al., 2024) explore teacher-agent interactions, few examine how these systems are implemented in authentic educational environments or how they affect teacher roles and practices. Research is needed on how teachers perceive, adopt, and adapt these agents within their existing pedagogical approaches, particularly regarding the balance of authority and responsibility between human and AI educators. Moreover, the literature shows limited theoretical grounding in established learning sciences. While some implementations reference educational theories (Wei et al.'s 5E model, for instance), many lack explicit connections to cognitive, social, or constructivist learning frameworks. This theoretical gap limits the extent to which these systems can be designed to align with established principles of effective learning and instruction. Future research should more systematically integrate learning sciences with the development of LLM-based pedagogical agents. Finally, there is minimal research on how student characteristics influence interactions with these agents. Factors such as prior knowledge, learning preferences, cultural background, and digital literacy likely affect how students engage with and benefit from LLM-based agents, yet these variables are rarely examined or controlled for. Future studies should investigate how these individual differences moderate agent effectiveness and how agent design might be adapted to address diverse student needs. As Lee, Shin et al. (2024) suggested, future research should "include learners with different learner profiles... and examine their interactions with LLM-based chatbots" (*p*. 11).

*4.2 Implications for Educational Practice and Policy*

The emergence of LLM-based pedagogical agents has significant implications for educational stakeholders beyond researchers and developers. For classroom teachers, these technologies offer both opportunities and challenges. On the one hand, systems like *TeachTune* (Jin et al., 2024) and *SimPal* (Farhana et al., 2024) demonstrate how these agents can extend teacher capabilities, providing personalized support that would be impractical for a single human educator to offer. On the other hand, their implementation requires careful integration with existing pedagogical approaches and technological ecosystems. Teachers will need professional development focused not merely on technical operations but also on pedagogically sound implementation. For instance, teachers must understand how to select appropriate agents, customize them for specific learning objectives, and evaluate their effectiveness. Teacher education programs should begin incorporating AI literacy components to prepare future educators for this changing landscape.

For educational institutions, these technologies necessitate new infrastructure and governance considerations. Technical requirements for supporting advanced LLM-based systems may exceed current capabilities, particularly in K-12 settings with limited IT resources. Institutions will need to evaluate both the direct costs of implementing these systems and the indirect costs of training, support, and maintenance (Casheekar et al., 2024; M. Liu & M'hiri, 2024). Additionally, they will need to develop policies governing appropriate use cases, data management practices, and evaluation standards.

For educational policymakers, LLM-based pedagogical agents present novel regulatory challenges (Borgogno & Perrazzelli, 2025). Unlike traditional educational materials, these systems exhibit emergent behaviors that cannot be fully predicted or controlled in advance. This characteristic complicates conventional approval processes for educational resources and raises questions about appropriate oversight. Policymakers should develop adaptive regulatory frameworks that balance innovation with appropriate safeguards (El-Deeb et al., 2024). They should also consider how to address potential disparities in access to these technologies, ensuring that advanced educational AI does not become another vector for educational inequality (Yu et al., 2024).

For students, the proliferation of these agents will require new forms of digital literacy (Jiao et al., 2024; Kumar et al., 2024). Beyond basic technological competence, students will need to develop appropriate mental models of AI capabilities and limitations. For instance, they need to understand when and how to productively engage with these systems, critically evaluate their outputs, and maintain their own agency in the learning process. As Liu, Yao et al. (2024) observed in their study of PeerGPT, children readily anthropomorphized AI agents while sometimes placing excessive trust in their responses. Educational programs should help students develop a clear understanding of AI systems as tools with specific affordances and constraints rather than authoritative or infallible sources.

### *4.3 Ethical Considerations and Concerns*

The adoption of LLM-based pedagogical agents raises significant ethical considerations that intersect with each of our identified design dimensions. Regarding the interaction approach, proactive systems that monitor student behavior to initiate interventions present heightened privacy concerns compared to purely reactive systems. For instance, while Zhang et al.'s (2024) proactive classroom agents demonstrate potential benefits for engagement, their continuous monitoring capabilities raise questions about surveillance in educational settings and potential chilling effects on student expression (Penney, 2017). Future research must develop frameworks for balancing the benefits of proactive support with appropriate privacy safeguards, potentially incorporating transparent monitoring policies and student control over agent activation.

In the domain scope dimension, general-purpose agents may present greater risks of generating inaccurate or misleading educational content than domain-specific systems with carefully curated knowledge bases (Wang et al., 2024). This concern is particularly acute in subjects requiring high factual accuracy, such as science or medicine (Collins et al., 2024; Pfohl et al., 2024). As Collins et al. (2024) demonstrated with *AnatomyGPT*, grounding agents in verified educational resources can mitigate hallucination risks, but this approach requires substantial domain expertise and resource investment. Research is needed on scalable approaches to knowledge verification that can maintain accuracy while accommodating diverse subject areas.

Role complexity introduces ethical questions about appropriate boundaries between AI and human educational roles. Multi-role systems like *MAIC* (Yu et al., 2024) that incorporate teacher, assistant, and peer personas blur traditional distinctions between these roles, potentially conflating different forms of educational authority. System integration raises data security concerns that increase with the breadth of integration. Integrated agents like *NivTA* (Jia et al., 2024) potentially access extensive student data across multiple educational platforms, creating significant security and consent challenges. As Martínez-Araneda et al. (2023) acknowledged regarding *TutorBot+*, storing and analyzing student interactions can provide valuable insights. However, it simultaneously raised data protection questions. Future implementations must develop transparent data governance frameworks that clearly communicate what student information is collected, how it is used, and how long it is retained (Gan et al., 2024).

Beyond these dimension-specific concerns, broader ethical questions emerge about student autonomy and critical thinking. While Bassner et al. (2024) found that students using *Iris* maintained confidence in their independent programming capabilities, the risk remains that agent assistance might undermine the development of self-directed learning skills. This concern is particularly relevant for younger students whose metacognitive abilities are still developing (Spivack et al., 2024). Future research should examine how different agent designs impact student self-efficacy and autonomous learning, potentially developing adaptive support models that gradually reduce assistance as student capabilities grow.

**5. Conclusion**

In this study, we conducted a scoping review of LLM-based pedagogical agents, a rapidly evolving field that is transforming educational technology through unprecedented natural language capabilities. Our analysis of 52 studies identified diverse applications across educational contexts, characterized by four key design dimensions: interaction approach (reactive vs. proactive), domain scope (domain-specific vs. general-purpose), role complexity (single-role vs. multi-role), and system integration (standalone vs. integrated). Moreover, we identified several directions for future research, including multi-agent systems for realistic educational environments, virtual student simulations for agent evaluation, integration with immersive technologies, multimodal interactions, and combining pedagogical agents with learning analytics. Furthermore, developing robust evaluation frameworks specifically designed for LLM-based pedagogical agents will be crucial for assessing both technical performance and educational effectiveness across different student populations and learning contexts. Addressing these research directions while considering ethical implications related to privacy, accuracy, and student autonomy will be essential for realizing the full potential of these emerging technologies in educational practice.

**Declaration of generative AI and AI-assisted technologies in the writing process**: All original thoughts, analyses, data interpretation, and conceptual frameworks were developed entirely by human authors. During the preparation of this work the author(s) used Claude to improve the readability and language of the manuscript. After using this tool/service, the author(s) reviewed and edited the content as needed and take(s) full responsibility for the content of the published article.

Appendix *Representative LLM-based Pedagogical Agents*

| **Study** | **Agent Name** | **Purpose/Goal** | **Role & Interaction** | **Context** | **Key Design Features** |
|---|---|---|---|---|---|
| Abdelghani et al. (2024) | Kids Ask (conversational agents for curiosity-driven learning) | To train children's curious question-asking skills and divergent questioning abilities, focusing on developing epistemic curiosity through interactive conversations | Functions as both an evaluator and tutor with two main workspaces:<br>• Workspace 1: Elicits curiosity through general knowledge quizzes and self-reflection activities<br>• Workspace 2: Trains divergent question asking through semantic and linguistic cues that guide students to formulate higher-level questions | Domain-general educational tool tested with middle school science content but designed to be adaptable across subjects | • State machine-based conversation management that enables adaptive dialogue paths through nodes and edges<br>• Personalized Reflect-Respond pipeline that integrates both knowledge states and student traits into LLM-simulated behaviors<br>• Dual-type cueing system combining structured "closed" cues for targeted questions and flexible "open" cues for exploratory learning |
| Bassner et al. (2024) | Iris | To provide personalized, context-aware assistance to computer science students working on programming exercises | Functions as a virtual tutor within the Artemis learning platform, offering subtle hints, counter-questions, and best practices rather than direct solutions | Designed for large-scale computer science education, particularly introductory programming courses (CS1) | • Context-awareness through automated access to exercise problem statements and student code repositories<br>• Few-shot learning with examples of appropriate responses to guide the LLM behavior<br>• Four-step interaction strategy: relevance assessment, file selection, response generation, and post-generation self-check |
| Chiang et al. (2024) | Python Pedagogical Agent | To support process-outcome-balanced educational practice in an introductory Python | Functions as an expert programming educator that guides students through coding exercises, evaluates submissions | Introductory Python programming course for non-IT university | • Integration of experiential learning cycle with customizable GPT-4 configuration |
| | | programming course by implementing an experiential learning cycle with ChatGPT | with detailed scoring rubrics, offers tailored hints without direct answers, poses reflective questions to deepen learning, and facilitates concept mapping. The agent documents their entire learning journey as portfolio evidence. | students, designed to accommodate diverse academic backgrounds and programming experience levels | • Pre-configured knowledge base containing weekly exercises and assessment questions<br>• Automated scoring system evaluating syntax, readability, and user-friendliness<br>• Basic Python programming knowledge map for concept visualization<br>• Capability to document entire learning process as portfolio evidence |
| Dan et al. (2023) | EduChat | To provide personalized, fair, and compassionate intelligent education support for teachers, students, and parents through a large language model-based chatbot system | Functions as an educational assistant with multiple roles including open question answering, essay assessment, Socratic teaching, and emotional support counselor. Interaction style adapts based on the specific educational function being used. | Designed for broad educational use, with specific focus on addressing challenges in applying LLMs to education | • Pre-training on extensive educational materials including textbooks, psychology texts, and instruction data<br>• Retrieval-augmented technology for accessing current information<br>• Integration of psychological and educational theories<br>• Multi-task conversation system with separate prompts for different educational tasks |
| Farhana et al. (2024) | SimPal (Meta-conversational physics education agent) | To assist teachers in adopting and customizing existing physics simulations for their lesson plans by helping them design instructional goals and customize | Functions as a meta-conversational assistant that engages teachers in natural dialogue about their instructional goals, extracts relevant physical variables and relationships, creates | Designed for K-12 physics education, specifically tested with 63 physics simulations from PhET and | • Natural language meta-conversation interface for teachers to explain their instructional goals<br>• Variable extraction system that identifies relevant physical concepts and relationships |
| | | a conversational AI agent that aligns with those goals through natural meta-conversation | symbolic representations, and helps design prompts for the original AI agent | Golabz platforms | • Symbolic representation generation to encode instructional goals<br>• Integration with multiple LLMs (ChatGPT-3.5 and PaLM 2) for processing teacher input |
| Hu et al. (2025) | CollaBot | To enhance students' engagement in online collaborative writing (OCW) activities by providing adaptive scaffolding support for groups working together on writing tasks | Functions as an adaptive support system with two activation mechanisms:<br>(1) Event-triggered: Monitors group behaviors and automatically provides scaffolding when needed<br>(2) User-initiated: Responds to student queries with personalized assistance | Designed for higher education settings, specifically collaborative essay writing tasks. The agent was integrated into a web-based writing platform. | • Integration of multiple AI technologies: retrieval-based model, generative AI model, and retrieval-augmented generation<br>• Three types of scaffolding: cognitive, metacognitive, and social scaffolding<br>• Two-module architecture: diagnosing module (analyzes student behaviors and text) and scaffolding module (provides appropriate support) |
| Jia et al. (2024) | NivTA (Naturally Interactable Virtual Teaching Assistant) | To create an interactive virtual teaching assistant for educational metaverse environments that can interact naturally with students | Functions as a life-sized virtual teaching assistant that students can engage with through natural interaction modes including speech, gestures, gaze, and spatial positioning | Deployed in a CAVE virtual environment, a visualization system for educational knowledge graphs primarily targeting higher education | • Integration of large language models (GPT-3.5) with natural interaction modalities<br>• Knowledge graph-based educational context feeding into LLM prompts<br>• Virtual avatar with human-like behaviors<br>• Motion tracking for detecting user gestures and position |
| Jin et al. (2024) | TeachTune (LLM-based pedagogical | To help teachers efficiently review and improve their | Functions as an evaluation tool that generates simulated student-agent | Used in K-12 science education but | • Personalized simulation pipeline that models both student |

| | agent review system) | pedagogical conversational agents by utilizing automated chats between the agent and simulated students with diverse profiles | conversations, allowing teachers to observe how their pedagogical agents adapt to different student types and scenarios | designed to be adaptable across subjects and grade levels | knowledge and traits (motivation, self-efficacy, stress)<br>• Graph-based interface for teachers to design agent conversation flows<br>• Automated chat generation between simulated students and pedagogical agents |
|---|---|---|---|---|---|
| Liu, Lu et al. (2024) | ScienceChat | To facilitate high school science video learning by providing LLM-empowered chat-based and quiz-based tutoring | Functions as both a chat tutor engaging in open-ended dialogue and a traditional quiz system providing multiple-choice questions with feedback. In Chat Mode, an AI agent guides students through discussions with follow-up questions and hints. In Quiz Mode, students receive immediate feedback on multiple-choice selections. | Designed for high school science education (grades 9-12), tested with videos on mutations and geoengineering topics. | • Retrieval-augmented generation to ground responses in video content<br>• Video timestamp-linked questions and hints<br>• Automatic conversion of open-ended questions to multiple-choice format<br>• Teacher interface for content creation and system configuration |
| Liu, Yao et al. (2024) | PeerGPT | To explore and assess the effectiveness of LLM-based peer agents in two distinct roles (team moderator and team participant) during children's | Functions in two roles:<br>• As team moderator: Facilitates discussion processes, guides transitions between phases, and manages team activities<br>• As team participant: Acts as a team member, engaging in discussions | Designed and tested in a collaborative design workshop environment with elementary school students (ages 11-12), but adaptable | • Built using GPT-3.5 with remote voice output devices<br>• Role-specific instruction sets for moderator and participant functions<br>• Customized prompting system for different roles |
| | | collaborative learning activities | and providing technical knowledge | for various collaborative learning scenarios across subjects and age groups. | |
| Liu, Zhu et al. (2024) | ClassMeta | To promote classroom participation in virtual reality classrooms by simulating active student behaviors and exerting conducive peer influence through GPT-4 powered agents | Functions as an active student who interacts with both instructors and peer students through spoken language and body gestures. Key interactions include answering/asking questions, taking notes, reminding about missing key points, promoting discussions, and correcting distracted behaviors. | Virtual reality classroom environment for educational institutions, tested with engineering design methodology content. Can be applied across various subjects and learning contexts. | • GPT-4 powered agents that process lecture materials as background context and classroom conversations as real-time context<br>• Uses 3D avatars with prerecorded human movements for authentic body language<br>• Integrates speech-to-text conversion, real-time voice synthesis, and attention tracking<br>• Includes a template system for educators to implement customized agents |
| Martin et al. (2024) | ChEdBot (Conversational agent for educational simulation) | To support simulation-based learning by enabling students to interview AI-powered domain experts and collect requirements, helping them develop interviewing skills | Functions as a domain expert that students can freely interview in a user-driven conversation style, rather than following predefined conversation paths | Domain-specific implementation for business education, focusing on requirements gathering and stakeholder interviews | • Frame-based dialogue system for managing structured conversations<br>• Customizable prompt templates for domain-specific knowledge integration<br>• Multi-agent system with distinct stakeholder personas and roles<br>• Integration of LLMs for generating natural, contextually-appropriate responses |
| Martínez-Araneda et al. (2023) | tutorBot+ | To provide effective, positive, and timely feedback to programming course students | Functions in two versions: Non-conversational version integrates with the university platform to provide automated code feedback when students submit solutions. Conversational version engages in natural dialogue, offering assistance for both problem analysis and code feedback for various programming languages | Designed for university-level computer science programming courses | • ChatGPT 3.5-powered feedback generation with carefully designed prompts<br>• Identification of strengths and weaknesses in student code<br>• Support for the ADCP (Analysis, Design, Construction, Testing) learning methodology |
| Matelsky et al. (2023) | FreeText | To automate assessment and feedback for open-ended student responses while reducing instructor workload and maintaining pedagogical quality | Functions as an automated grading assistant that evaluates student responses against instructor-defined criteria | Designed for various educational environments, particularly useful in large classes where personalized feedback is time-intensive | • Instructor-defined grading criteria and feedback templates<br>• LLM-powered automated response evaluation<br>• Privacy-preserving design that maintains information asymmetry between students and system<br>• Support for both standalone operation and integration with existing educational platforms |
| Pan et al. (2024) | ELLMA-T (Embodied Large Language Model Agent for Tutoring) | To support English language learning for adult learners by providing immersive, personalized, and contextual language practice in social | Functions as a dynamic language tutor that dynamically adapts its interaction to each learner's needs. Specifically, the agent acts as a conversational partner, assessment | Targeted adult learners for informal English language learning | • Embodied avatar in virtual environment<br>• Adaptive language difficulty adjustment<br>• Scenario-based role-play conversations |
| | | virtual reality environments | facilitator, and supportive learning guide | | • Multi-modal interaction supporting verbal and non-verbal communication |
| Teng et al. (2024) | CVTutor | To enhance interactive learning and student outcomes in flipped classroom environments | Comprising three specialized sub-agents: QA_TA, GraderTA, and EthicsTA. QA_TA answers student questions and stimulates engagement; GraderTA assesses assignments and provides feedback; EthicsTA monitors content for ethical compliance. | Implemented in a 6-week undergraduate computer vision course | • Retrieval-augmented generation (RAG) technology to access domain-specific knowledge<br>• Multi-agent collaborative framework with feedback loops for continuous improvement<br>• Integration with various tools including Whisper for speech-to-text conversion |
| Volkmann (2024) | EduBot | To measure and enhance digital competencies in online collaborative learning environments through data-driven analysis of user activities that trigger self-reflection for competence development | EduBot collects and analyzes student activity data to provide personalized feedback on digital behaviors, focusing on communication patterns, ethical practices, and collaborative tool use. It interacts with students through a chat interface, delivering weekly individualized feedback reports. | Implemented in a Virtual Collaborative Learning (VCL) framework with 20 university students in an eight-week module combining blended learning and VCL | • Integration of Learning Analytics (LA) and LLMs to analyze student activity data<br>• Natural language-based analytical interface<br>• Visual dashboard displaying communication patterns and activity distribution<br>• Automated, semi-adaptive notifications |
| Wang et al. (2024) | Multi-Agent MCQ Generation System | To address the lack of scalable and reliable AI literacy assessment materials | A multi-agent workflow that collaboratively generates and refines AI literacy assessment | Focused on creating AI literacy assessment | • Incorporates multiple agents with specific roles (Generator Agent produces initial questions; Language Critique Agent |
| | | by developing an automated system for generating multiple-choice questions (MCQs) using large language models | questions through iterative review and critique | materials for K-7-9 students, addressing the educational need for AI literacy resources | evaluates readability and grade-level appropriateness; Item-Writing Flaw Critique Agent checks for potential question composition issues; and Supervisor Agent approves or requests revisions. |
| Wei et al. (2024) | Generative AI-empowered Pedagogical Agent (GPA) system | To support science education through the 5E learning model (Engagement, Exploration, Explanation, Elaboration, and Evaluation), helping students learn astronomy concepts | Multi-role system where:<br>• Instructor GPA delivers predetermined content through AR scenes and 3D models<br>• Guide GPA facilitates learning activities and provides structured support<br>• Conversation GPA responds to student queries with contextual explanations | Domain-specific implementation focused on elementary science education, particularly astronomy concepts | • Integration with Unity's AR platform and Vuforia plugin for immersive learning<br>• Seamless connection with ChatGPT API for conversational capabilities<br>• Access to databases containing guiding content, teaching materials, and assessment resources |
| Xu et al. (2024) | EduAgent (Student behavior simulation system) | To simulate student learning behaviors and interactions to help evaluate pedagogical agents and learning content before deployment | Functions as simulated students who interact with teachers and tutoring systems, demonstrating realistic learning behaviors and cognitive states during conversations | Initially tested with middle school science topics but designed as a domain-general framework for simulating student behaviors | • Configurable student profiles combining academic and psychological characteristics<br>• Dynamic knowledge state tracking that simulates learning progress<br>• Module for simulating diverse cognitive states and student characteristics |
| Yu et al. (2024) | MAIC (Massive AI-empowered Course) | To transform traditional Massive Open Online Courses (MOOCs) into more adaptive and personalized learning experiences using large language model-driven multi-agent systems | Creates an AI-augmented classroom environment with multiple agents including a teacher agent, teaching assistant, and classmate agents who interact dynamically to support student learning | Developed and initially tested at Tsinghua University across two courses - an AI course and a learning skills course, involving over 500 student volunteers | • Multi-agent system with distinct roles like teacher, assistant, and classmate agents<br>• Agents capable of providing emotional support, maintaining classroom discipline, and facilitating discussions<br>• Ability to dynamically adjust teaching based on student interactions and needs<br>• Scalable approach balancing educational accessibility with personalized learning experiences |
| Zhang et al. (2024) | SimClass (Classroom Simulation Framework) | To develop a multi-agent classroom simulation system that can create dynamic, interactive learning environments using large language models | Creates a simulated classroom with multiple agents representing different roles (teacher, teaching assistant, classmate agents) who interact to facilitate learning and simulate realistic educational experiences | Tested primarily in science education contexts, focusing on creating more engaging and adaptive online learning interactions | • Agents with distinct pedagogical roles and personalities<br>• A session controller managing conversation flow and agent interactions<br>• Ability to simulate teaching and initiation, in-depth discussion, emotional companionship, and classroom management |